\documentclass[conference]{IEEEtran}\IEEEoverridecommandlockouts
\usepackage{cite}
\usepackage{amsmath,amssymb,amsfonts}
\usepackage{algorithmic}
\usepackage{graphicx}
\usepackage{textcomp}
\usepackage{xcolor}
\usepackage{epsfig}
\usepackage{float}
\usepackage{bm}
\usepackage{graphicx}
\usepackage{amsfonts,amssymb}
\usepackage{booktabs}
\usepackage{array}
\usepackage{tabularx}
\usepackage{multirow}
\usepackage{subfigure}
\usepackage{color}
\usepackage{epstopdf}
\usepackage{makecell}
\usepackage{url}
\usepackage{gensymb}
\usepackage{amsmath}
\usepackage{xcolor}
\usepackage[implicit=false]{hyperref}

\usepackage[ruled]{algorithm2e}
\def\BibTeX{{\rm B\kern-.05em{\sc i\kern-.025em b}\kern-.08em
	T\kern-.1667em\lower.7ex\hbox{E}\kern-.125emX}}

\begin{document}

\title{Luminance-Guided Chrominance Image Enhancement for HEVC Intra Coding\vspace{-.5em}}

\author{\IEEEauthorblockN{Hewei Liu$^{\ast}$, Renwei Yang$^{\ast}$, Shuyuan Zhu$^{\ast}$, Xing Wen$^{\dag}$ and Bing Zeng$^{\ast}$}
	\IEEEauthorblockA{$^{\ast}$School of Information and Communication Engineering\\
		University of Electronic Science and Technology of China, Chengdu, China\\
		$^{\dag}$Kuaishou, Beijing, China\\}

\thanks{This work was supported by the National Natural Science Foundation of
        	China under Grant U20A20184 and Grant 62031009. The first and second authors contributed equally to this work.}
}

\maketitle

\begin{abstract}
	In this paper, we propose a luminance-guided chrominance image enhancement convolutional neural network for HEVC intra coding. Specifically, we firstly develop a gated recursive asymmetric-convolution block to restore each degraded chrominance image, which generates an intermediate output. Then, guided by the luminance image, the quality of this intermediate output is further improved, which finally produces the high-quality chrominance image. When our proposed method is adopted in the compression of color images with HEVC intra coding, it achieves 28.96\% and 16.74\% BD-rate gains over HEVC for the U and V images, respectively, which accordingly demonstrate its superiority. The code is available online: https://github.com/Nickyang4900/Luminance-Guided-Chrominance-Enhancement-for-HEVC-Intra-Coding.
\end{abstract}

\vspace{.5em}

\section{Introduction}

With the rapid development of multimedia technology, there has been a huge demand for high-definition content. Currently, high efficiency video coding (HEVC) \cite{hevc} is widely used in the compression of high-definition content. However, still images compressed by HEVC intra coding often suffer from the compression artifacts, especially at the low bit-rate. Recently, the new generation of video coding standard, i.e., versatile video coding (VVC) \cite{vvc} has been proposed. VVC offers more excellent rate-distortion performance compared with HEVC. However, the high computational complexity of VVC limits its application in practice. Improving the performance of HEVC is still very necessary and challenged. Enhancing the quality of compressed image offers a simple but effective way to obtain a high coding efficiency. Over the past few years, the convolutional neural network (CNN) based solution is becoming more and more popular to achieve this goal.

The CNN models are normally used to compose the end-to-end schemes to reduce the compression artifacts \cite{jpeg1,jpeg2}. In \cite{saocnn}, the CNN-based in-loop filter was designed to replace the sample adaptive offset (SAO) \cite{sao} in HEVC. In \cite{dbcnn}, a multi-channel long-short-term dependency residual network was proposed to enhance the quality of video frames based on the temporal correlation of frames. The CNN-based post-processing methods were proposed  in \cite{vrcnn} and \cite{post2} for the HEVC-coded images. More specifically, a variable-filter-size residue-learning was designed in \cite{vrcnn} to implement the quality enhancement and a deep CNN-based auto decoder was proposed in \cite{post2} to achieve the same goal. In \cite{Jia2019}, several enhancement networks were designed to improve the quality of different coding tree units which finally achieves a overall quality gain for the whole image. Meanwhile, the coding information was also adopted in the CNN-based quality enhancement solution. For example, in \cite{Ma2019}, the motion residue was utilized to improve the video quality. The coding unit mask was defined in \cite{cumask} as the prior information for the CNN-based post-processing. Also, the transform unit partition map was employed in  \cite{Meng2020} to improve the quality of video frames.

\begin{figure}[t]
	\centering
	\includegraphics[height=4.5cm,width=9cm]{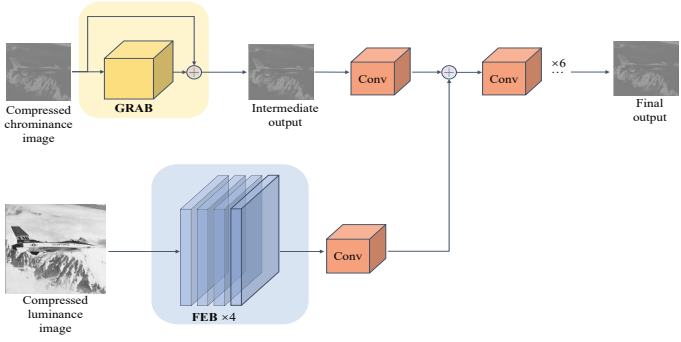}
	\small{\caption{Pipeline of our proposed method.}\label{1-overview}}
\end{figure}

Besides the coding information, the channel correlation between the luminance image (Y) and the chrominance images (U and V) may also be utilized to improve the quality of compressed images. However, most of the existing CNN-based methods process each channel independently, which ignores the channel correlation and cannot achieve a high efficiency. In this work, we propose  a luminance-guided chrominance image enhancement method for HEVC intra coding. More specifically, the degraded chrominance image is firstly fed into the recursive asymmetric-convolution block (GRAB) to obtain an intermediate output. Then, the feature of luminance image is obtained via the feature extraction block (FEB). Finally, the intermediate output and the luminance feature are combined together to reconstruct the final chrominance images. The pipeline of our proposed method is illustrated in Fig. \ref{1-overview}.

The rest of this paper is organized as follows. The motivation of our proposed method is given in Section \ref{Mot}. Our proposed method is introduced in Section \ref{Ours}. The experimental results are presented in Section \ref{Results} and the conclusions are drawn in Section \ref{Con}.

\begin{figure*} [t]
	\centering
	\includegraphics[height=3.5cm,width=17.5cm]{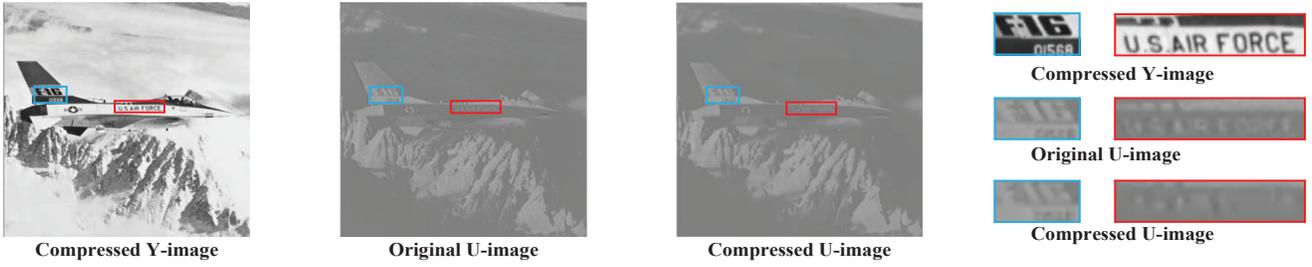}
	
	\small{\caption{An example of YUV-420 format coding by HEVC intra coding, where the quantization parameter is 22.}\label{yuv-relationship}}
\end{figure*}

\begin{figure*} [t]
	\centering
	\includegraphics[height=3.5cm,width=14cm]{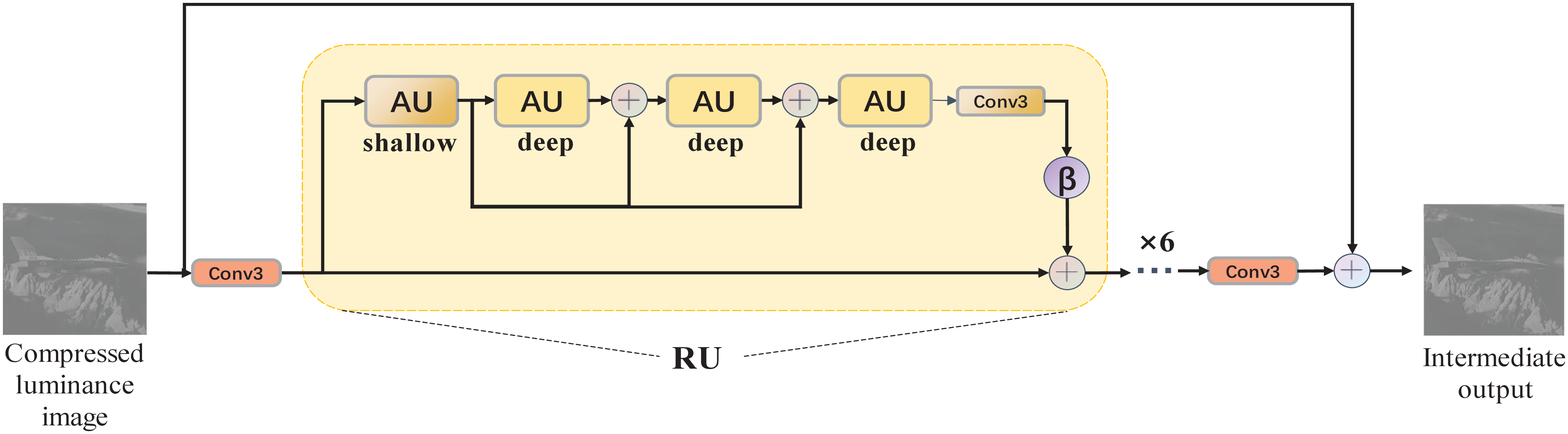}
	\small{\caption{Architectures of GRAB.}\label{net1}}
\end{figure*}

\section{Motivation} \label{Mot}

In color image compression, the input RGB image is normally converted into the YUV format, including one luminance image (Y) and two chrominance images, i.e., U and V. Both the U and V images are then spatially down-sampled to form the YUV-420 format. The RGB-YUV conversion is implemented as

\begin{equation}\label{Eq.yuv}
\scalebox{.95}{$
	\!\begin{bmatrix}
	Y\\U\\V
	\end{bmatrix}\!=\!
	\begin{bmatrix}\!
	0.2126&0.7152&0.0722\\-0.1146&-0.3854&0.5000\\0.5000&-0.4542&-0.0458
	\!\end{bmatrix}\! \!
	\begin{bmatrix}\!
	R\\G\\B
	\!\end{bmatrix}\!+\begin{bmatrix}\!
	16\\128\\128
	\!\end{bmatrix}\!.$}
\end{equation}

According to Eq. (\ref{Eq.yuv}), although the luminance and chrominance images are obtained by combining the red (R), green (G) and blue (B) channels with different weighting factors, they are still highly correlated. We present some YUV-420 coding results of HEVC intra coding in Fig. \ref{yuv-relationship}. It is found from Fig. \ref{yuv-relationship} that the compressed luminance image may potentially offer some detailed textures to enhance the compressed chrominance image. More specifically, the content of the luminance image is much clear but the texture of the chrominance image is blurred due to the spatial down-sampling and compression. The clear luminance textures may potentially provide prior information to enhance the quality of the chrominance image, which will finally offer an overall quality gain for the whole image.

\section{Our Proposed Method} \label{Ours}

\subsection{Overview}

Our proposed method is designed for the YUV-420 format coding after the RGB-YUV conversion. Its pipeline is illustrated in Fig. \ref{1-overview}, where the decoded chrominance image is firstly enhanced by GRAB, and then is further enhanced with the guidance of the decoded luminance image. We stack four feature extraction blocks (FEB) to extract the prior information from the luminance image. In our work, the intermediate chrominance output is passed through a $3\times3$ convolution kernel with stride = 1. Therefore, we adopt the $3\times3$ convolution kernel with stride = 2 after four FEBs to guarantee the luminance features and the chrominance features with the same size. To fuse the guided features offered by the luminance image, the features of the intermediate chrominance output and the ones of the luminance image are combined together via the element-wise adding. Moreover, we employ six convolution layers with the $3 \times 3$ kernel size and stride = 1 in the last reconstruction procedure to yield the final chrominance output.

\subsection{Gated Recursive Asymmetric-convolution Block}\label{3.1}

\begin{figure}[t]
	\centering
	\includegraphics[height=1.5cm,width=8cm]{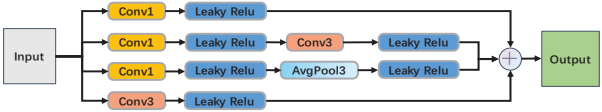}
	\small{\caption{Architectures of AU.}\label{net2}}
\end{figure}

The proposed GRAB is used to produce the intermediate output for the chrominance image, and it is composed by six recursive units (RU). The recursive design adopted in our work aims at avoiding the dramatic increasing of model parameters when the network depth increases \cite{Kim2016}. The architecture of GRAB is given in Fig. \ref{net1}. In GRAB, each RU consists of four asymmetric units (AU) \cite{Ding2021}, including a shallow one and three deep ones, a convolution layer, and a gate coefficient (GC) \cite{Tai2017}. Due to the adoption of skip connection \cite{He2015}, RU only needs to learn the residual between the source high-quality image and the compressed low-quality input image, which leads to a simple training\cite{Zhang2017}. Except for the given values of GC, the six RUs share the same architecture and weighting parameters. In each RU, AU is designed with various branches, thus it is able to strengthen the network capability to obtain diverse features from input \cite{Ding2019,Ding2021}. Each AU consists of three $1\times1$ convolution layers, two $3\times3$ convolution layers, one average pooling and Leaky ReLU. The structure of AU is given in Fig. \ref{net2}, and it is implemented as

\vspace{-.5em}
\begin{equation}\label{Eq.AU}
z = c_{1}(x) + c_{3}(c_{1}(x)) + AvgPool_{3}(c_{1}(x)) + c_{3}(x)
\end{equation}

\noindent
where $c_{i\in \{1,3\}}(\cdot)$ represents the convolution operation with the $i \times i$ kernel size, and $AvgPool_{3}$ denotes the average pooling with the 3$\times$3 slide window. In our work, the shallow and deep AUs share the same structures, but employ different weighting parameters, which can effectively reduce the parameters.

The gate coefficient was initially proposed in \cite{Tai2017} to solve the image super resolution problem. We adopt it in this work to construct the depth-wise attention mechanism so that the network can be trained to assign different weights to the features of different depths, which potentially leads to a good reconstruction result.  Moreover, we do not employ the channel attention methods \cite{Zhang2018,Hu2019} in our work to avoid the high computational cost. The gate coefficient $\beta$ adopted in GRAB is used as the weighting factor for each RU and it is achieved via multiplication as

\vspace{-1em}
\begin{eqnarray} \label{Eq.Gate}
z = \beta \cdot x.
\end{eqnarray}

Let us use $s(\cdot)$ and $d(\cdot)$ to represent the operation of the shallow and deep AUs, respectively. Then, in the $i$-th RU, we can obtain the output of the last AU

\begin{equation}
AU_4(y_{i-1}) =  d(d(d(s(y_{i-1}))+ s(y_{i-1}))+s(y_{i-1}))
\end{equation}

\noindent
where $i \in\{1,\ldots,6\}$, and $y_{i}$ and $\beta_{i}$ denote the output and GC of the $i$-th RU, respectively. Note that $y_0$ is the output obtained by inputting the compressed chrominance image to a convolution layer whose kernel size is $3 \times 3 $. Finally, the procedure of passing through the $i$-th RU is formulated as

\begin{eqnarray}\label{Eq.cbenhance1}
\begin{array}{l}
\mathop y_{i} = \beta_{i} \cdot conv(AU(y_{i-1})) + y_{i-1}.\\
\end{array}
\end{eqnarray}

\subsection{Feature Extraction Block}\label{3.2}

FEU is designed to extract features of the luminance image and it is designed with the skip connection structure, allowing the combination of shallow and deep convolution layer outputs. This structure not only strengthens the ability to extract features, but also makes the training more easily \cite{Kim2016}. The architecture of the FEB is illustrated in Fig. \ref{net3}. In this unit, the feature maps of all the depths directly connect to the last layer, which combines the shallow and deep features from the input image. In our work, FEB is implemented as

\begin{equation}\label{Eq.cbenhance1}
f_{FEB} = \sum_{i=1}^4 {\gamma}_{i}
\end{equation}

\noindent
where ${\gamma}_i$ is the output of the $i$-th convolution layer, i.e., {${\gamma}_i=ReLU(W_i \ast {\gamma}_{i-1})$, $\ast$ denotes the convolution operation and $W_{i}$ is the weighting parameter of the convolution layer. In this work, the ReLU activation function $ReLU$ is employed to obtain ${\gamma}_i$. Moreover, ${\gamma}_0$ should be the input image for the first FEB in in our proposed network.

	\begin{figure}[t]
		\centering
		\includegraphics[height=1.5cm,width=8.5cm]{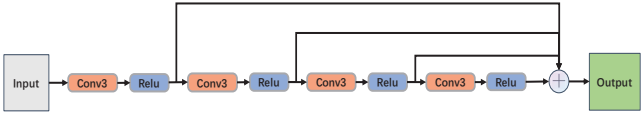}
		\small{\caption{Architectures of FEB.}\label{net3}}
	\end{figure}
	
	\begin{figure}[t]
		\centering
		\includegraphics[height=1cm,width=8.5cm]{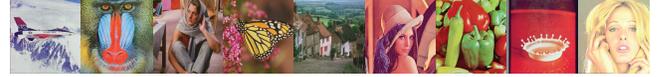}
		\small{\caption{Classical image dataset.}\label{classical}}
	\end{figure}
	
	\begin{table*}[t]
		\renewcommand\arraystretch{1.2}
		\centering
		\fontsize{6.5}{9}\selectfont
		\caption{Comparison of BD-rate.}\label{res1}
		\vspace{-.5em}{
			\begin{tabular} {c c c c c c c c c c}
				\Xhline{1.2pt}
				\hline\hline
				\multicolumn{2}{c}{}
				&\multicolumn{2}{c}{ASN-1}
				&\multicolumn{2}{c}{Ours-1}
				& \multicolumn{2}{c}{ASN-2}
				& \multicolumn{2}{c}{Ours-2}\\
				\cline{3-10}
				\multicolumn{2}{c}{} &U &V &U &V &U &V &U &V\\
				\hline
				
				&Classical &-9.06\%  & -8.91\%  &-9.60\%  &-9.68\%  & \textbf{-22.38\%}  &-11.39\% &-21.92\%  &\textbf{-14.33\%} \\
				&McMater   &-9.77\%  &-12.01\% &-10.23\%   &-12.28\%  &-30.02\%  &-15.51\%  &\textbf{-30.99\%} &\textbf{-17.56\%}   \\
				&Kodak     &-14.67\%  &-15.07\% & -15.23\% &-15.47\%  &-33.21\%  &-17.51\%  &\textbf{-33.98\%} &\textbf{-18.33\%}   \\
				&\textbf{Average}  &-11.17\% &-12.00\% &-11.68\%&-12.47\% &-28.54\%&-14.80\% &\textbf{-28.96\%} &\textbf{-16.74\%}
				\\\hline\hline
				\Xhline{1.2pt}
		\end{tabular}}
		\label{tab1}
		\vspace{-1em}
	\end{table*}

	\section{Experimental Results} \label{Results}
	
	\subsection{Experiment Setup}

	We verify the effectiveness of our proposed method by employing it to process the HEVC-compressed color images, where the HM 16.7 \cite{hm} is adopted in this work. Moreover, we compare the performance of our proposed method with the other existing methods. The first 800 color images of Flickr2K dataset \cite{flickr2k} are selected as the training dataset. Moreover, these images are converted into YUV-420 format and compressed by HEVC intra coding with the quantization parameters (QP) 22, 27, 32 and 37. The compressed chrominance images, including both the U and V images, are cropped into the $32 \times 32$ non-overlapping sub-images, and the compressed luminance images are cropped into the $64 \times 64$ non-overlapping sub-images. We train all the CNN models by using these generated sub-images.
	\begin{table}[t]
		\renewcommand\arraystretch{1.2}
		\centering
		\fontsize{6.5}{9}\selectfont
		\caption{Comparison of $\Delta$PSNR (dB) at QP 22.}\label{22}
		\vspace{-.5em}{
			\begin{tabular} {c c c c c c c c c c}
				\Xhline{1.2pt}
				\hline\hline
				\multicolumn{1}{c}{}
				&\multicolumn{2}{c}{ASN-1}
				&\multicolumn{2}{c}{Ours-1}
				& \multicolumn{2}{c}{ASN-2}
				& \multicolumn{2}{c}{Ours-2}\\
				\cline{2-9}
				\multicolumn{1}{c}{} &U &V &U &V &U &V &U &V\\
				\hline
				Classical
				&0.191	 &0.164	 &0.187	 &0.164	 &0.466	 &0.133	 &\textbf{0.491}	 &\textbf{0.203}		\\
				McMater
				&0.244	 &0.364	 &0.237	 &0.349	 &0.860	 &0.322	 &\textbf{0.940}	 &\textbf{0.427}  \\
				Kodak
				&0.371	 &0.460	 &0.365	 &0.441	 &0.919	 &0.455	 &\textbf{1.035}	 &\textbf{0.498}\\
				\textbf{Average}
				&0.269	 &0.329	 &0.263	 &0.318 &0.748	 &0.303	 &\textbf{0.822} &\textbf{0.376}
				\\\hline\hline
				\Xhline{1.2pt}
		\end{tabular}}
		\vspace{-1em}
	\end{table}
	
	\begin{table}[t]
		\renewcommand\arraystretch{1.2}
		\centering
		\fontsize{6.5}{9}\selectfont
		\caption{Comparison of $\Delta$PSNR (dB) at QP 27.}\label{27}
		\vspace{-.5em}{
			\begin{tabular} {c c c c c c c c c c}
				\Xhline{1.2pt}
				\hline\hline
				\multicolumn{1}{c}{}
				&\multicolumn{2}{c}{ASN-1}
				&\multicolumn{2}{c}{Ours-1}
				& \multicolumn{2}{c}{ASN-2}
				& \multicolumn{2}{c}{Ours-2}\\
				\cline{2-9}
				\multicolumn{1}{c}{} &U &V &U &V &U &V &U &V\\
				\hline
				Classical
				&0.304&0.269	 &0.319&0.297	 &\textbf{0.791}&0.325	 &0.781&\textbf{0.444}	 \\
				McMater
				&0.388&0.516	 &0.401&0.527	 &1.369&0.653	 &\textbf{1.430}&\textbf{0.752}	 \\
				Kodak
				&0.515&0.567	 &0.530&0.582	 &1.304&0.632	 &\textbf{1.367}&\textbf{0.688}	 \\
				\textbf{Average}
				&0.402&0.450	 &0.417&0.469	 &1.155&0.537	 &\textbf{1.193}&\textbf{0.628}
				
				\\\hline\hline
				\Xhline{1.2pt}
		\end{tabular}}
		\vspace{-1em}
	\end{table}
	
	\begin{table}[t]
		\renewcommand\arraystretch{1.2}
		\centering
		\fontsize{6.5}{9}\selectfont
		\caption{Comparison of $\Delta$PSNR (dB) at QP 32.}\label{32}
		\vspace{-.5em}{
			\begin{tabular} {c c c c c c c c c c}
				\Xhline{1.2pt}
				\hline\hline
				\multicolumn{1}{c}{}
				&\multicolumn{2}{c}{ASN-1}
				&\multicolumn{2}{c}{Ours-1}
				&\multicolumn{2}{c}{ASN-2}
				&\multicolumn{2}{c}{Ours-2}\\
				\cline{2-9}
				\multicolumn{1}{c}{} &U &V &U &V &U &V &U &V\\
				\hline
				Classical
				&0.337&0.360	 &0.360&0.384	 &\textbf{0.948}&0.509	 &0.909&\textbf{0.623}	 \\
				McMater
				&0.469&0.568	 &0.503&0.589	 &1.641&0.802	 &\textbf{1.692}&\textbf{0.908}	 \\
				Kodak
				&0.571&0.619	 &0.606&0.645	 &1.506&0.784	 &\textbf{1.528}&\textbf{0.797}	 \\
				\textbf{Average}
				&0.459&0.515	 &0.490&0.539	 &1.365&0.699	 &\textbf{1.376}&\textbf{0.776}
				
				\\\hline\hline
				\Xhline{1.2pt}
		\end{tabular}}
		\vspace{-1em}
	\end{table}
	
	\begin{table}[t]
		\vspace{0.5em}
		\renewcommand\arraystretch{1.2}
		\centering
		\fontsize{6.5}{9}\selectfont
		\caption{Comparison of $\Delta$PSNR (dB) at QP 37.}\label{37}
		\vspace
		{-1.35em}
		{
			\begin{tabular} {c c c c c c c c c c}
				\Xhline{1.2pt}
				\hline\hline
				\multicolumn{1}{c}{}
				&\multicolumn{2}{c}{ASN-1}
				&\multicolumn{2}{c}{Ours-1}
				& \multicolumn{2}{c}{ASN-2}
				& \multicolumn{2}{c}{Ours-2}\\
				\cline{2-9}
				\multicolumn{1}{c}{} &U &V &U &V &U &V &U &V\\
				\hline
				Classical
				&0.349&0.392	 &0.404&0.467	 &0.946&0.620	 &\textbf{0.947}&\textbf{0.718}	 \\
				McMater
				&0.493&0.589	 &0.522&0.595	 &1.576&0.828	 &\textbf{1.628}&\textbf{0.925}	 \\
				Kodak
				&0.585&0.674	 &0.598&0.685	 &1.401&0.775	 &\textbf{1.453}&\textbf{0.861}	 \\
				\textbf{Average}
				&0.476&0.552	 &0.508&0.582	 &1.308&0.741	 &\textbf{1.343}&\textbf{0.835}	
				\\\hline\hline
				\Xhline{1.2pt}
		\end{tabular}}
		\vspace{-1em}
	\end{table}
	
	\subsection{Implementation Detail}
	
	In the network training, the batch size is set to 32 and the total epoch is set to 40. We start with a learning rate of 10$^{-4}$ and decay the learning rate with 0.1 at the {20}th epochs. We first train the whole model at QP = 37 from scratch and the models at QP = 22, 27 and 32 are fine-tuned based on this trained model. Our proposed network is implemented based on Pytorch and optimized by the Adam optimizer via minimizing the $l_1$ loss.  All the models are trained with the NVIDIA GTX 2080TI GPU. Moreover, all models are only trained to enhance the quality of the U-image and they are also applied to improve the quality of the V-image.

	\subsection{Performance Comparison}
	
	To make a comprehensive comparison, three popular image datesets are utilized in our experiments, including 1) the classical image dataset, containing 9 color images as shown in Fig. \ref{classical}, 2) McMater dataset \cite{Mcm}, containing 18 color images, and 3) Kodak dataset \cite{Kodak}, containing 24 color images. These test datasets are compressed by HEVC at QP 22, 27, 32 and 37. The default HM is set as benchmark, and we test the quality enhancement performance in terms of BD-Rate and $\Delta$PSNR. The proposed approach is compared with adaptive-switching neural network (ASN) \cite{cumask} which also introduces the guidance image to enhance the quality of the HEVC-compressed images.
	
	We present the average BD-rate and average $\Delta$PSNR results over all the test images of each dataset in Tables \ref{tab1}-\ref{37}, respectively, where Ours-1 and Ours-2 represent our proposed method without and with the guidance of the Y-image, respectively. Similarly, we use  ASN-1 and ASN-2 to represent the ASN method without and with the guidance of the Y-image, respectively.
	
	Firstly, one can see from Table \ref{tab1} that our proposed method achieves 28.96\% (on average) BD-rate gain over HEVC for the U-images and 16.74\% (on average) gain for the V-images. Moreover, for most test image datasets, our approach offers higher BD-rate gain than ASN, not only for the processing of the U-images, but also for the processing of the V-images. Secondly, according to the $\Delta$PSNR results presented in Tables \ref{22}-\ref{37}, our proposed method does not perform as well as ASN when it is applied to process the chrominance images without the guidance image, especially at QP = 22. After the guidance image is introduced in our method, the significant performance gain is achieved. Meanwhile, when the images are compressed with larger QP, our method achieves higher $\Delta$PSNR (on average). Thirdly, all the results given in Tables \ref{tab1}-\ref{37} demonstrate that employing the luminance image to guide the enhancement of the chrominance images can provide remarkable gain for our proposed method.

	\section{Concluding Remarks} \label{Con}
	
	In this paper, we propose a luminance-guided chrominance image enhancement method for HEVC intra coding. We design the recursive asymmetric-convolution block to initially restore each degraded chrominance image. Then, the compressed luminance image is introduced as the guidance to further improve the quality of the chrominance image. The experimental results demonstrate that our proposed approach achieves 28.96\% and 16.74\% BD-rate gains over HEVC for the U- and V-images, respectively, when it is adopted in the compression of color images.

\end{document}